\documentclass[letterpaper]{article} % DO NOT CHANGE THIS
\usepackage{aaai2026}  % DO NOT CHANGE THIS (No [submission] option)
\usepackage{times}  % DO NOT CHANGE THIS
\usepackage{helvet}  % DO NOT CHANGE THIS
\usepackage{courier}  % DO NOT CHANGE THIS
\usepackage[hyphens]{url}  % DO NOT CHANGE THIS
\usepackage{graphicx} % DO NOT CHANGE THIS
\usepackage{natbib}  % DO NOT CHANGE THIS AND DO NOT ADD ANY OPTIONS TO IT
\usepackage{caption} % DO NOT CHANGE THIS AND DO NOT ADD ANY OPTIONS TO IT
\usepackage{newfloat}
\usepackage{listings}

\usepackage{booktabs}
\usepackage{amsmath}
\usepackage{amsfonts}
\usepackage{multirow}
\usepackage{ragged2e}
\usepackage{array}
\usepackage{microtype}

\title{Seeing Justice Clearly: Handwritten Legal Document Translation \\ with OCR and Vision-Language Models}

\author {
    Shubham Kumar Nigam\textsuperscript{\rm 1, 3}\equalcontrib\thanks{Corresponding Author},
    Parjanya Aditya Shukla\textsuperscript{\rm 1}\equalcontrib, 
    Noel Shallum\textsuperscript{\rm 2}, 
    Arnab Bhattacharya\textsuperscript{\rm 1}
}
\affiliations {
    \vspace{0.6em}
    \textsuperscript{\rm 1}Department of Computer Science, IIT Kanpur, Kanpur, India \quad
    \textsuperscript{\rm 2}Symbiosis Law School, Pune, India\\
    \textsuperscript{\rm 3}University of Birmingham, Dubai, UAE\\
    \vspace{0.8em}
    shubhamkumarnigam@gmail.com, \hspace{0.5em} padityashukla26@gmail.com, \hspace{0.5em} noelshallum@gmail.com, \hspace{0.5em} arnabb@cse.iitk.ac.in
}

\begin{document}

\maketitle

\begin{abstract}
Handwritten text recognition (HTR) and machine translation continue to pose significant challenges, particularly for low-resource languages like Marathi, which lack large digitized corpora and exhibit high variability in handwriting styles. The conventional approach to address this involves a two-stage pipeline: an OCR system extracts text from handwritten images, which is then translated into the target language using a machine translation model. In this work, we explore and compare the performance of traditional OCR-MT pipelines with Vision Large Language Models that aim to unify these stages and directly translate handwritten text images in a single, end-to-end step. Our motivation is grounded in the urgent need for scalable, accurate translation systems to digitize legal records such as FIRs, charge sheets, and witness statements in India's district and high courts. We evaluate both approaches on a curated dataset of handwritten Marathi legal documents, with the goal of enabling efficient legal document processing, even in low-resource environments. Our findings offer actionable insights toward building robust, edge-deployable solutions that enhance access to legal information for non-native speakers and legal professionals alike.
\end{abstract}

\begin{links}
\link{Code}
{https://github.com/anviksha-lab-iitk/SJC}
\end{links}

\section{Introduction}
\label{sec:introduction}
\begin{figure}[tb]
  \centering
  \includegraphics[width=\columnwidth]{figures/figure1.png}
  \caption{Comparison of OCR-MT and vLLM-based approaches for handwritten text translation. The OCR-MT pipeline decomposes the task into separate HTR and MT stages, whereas vLLMs unify the process into a single end-to-end step.}
  \label{fig:approaches}
\end{figure}

The Indian judiciary, one of the world's most complex legal systems, continues to face challenges in ensuring timely justice and efficient case handling. A major bottleneck lies in its persistent reliance on handwritten documentation at the grassroots level, such as district courts and police stations, where First Information Reports (FIRs), case diaries, witness statements, and court proceedings are still manually recorded. These documents are critical to criminal and civil proceedings, but their handwritten, unstructured nature makes them difficult to archive, search, and analyze. Variability in handwriting, language diversity, legal terminology, and the poor quality of scans all pose significant hurdles to digitization.

This work lays the foundation for building an automated legal document digitization pipeline aimed at enabling structured and accessible digital case records. Our focus is on automatically converting handwritten legal documents in Marathi into English through two distinct paradigms: (i) a modular OCR + Machine Translation (OCR-MT) pipeline, and (ii) direct translation using Vision Large Language Models (vLLMs). The objective is to assess their performance, robustness, and suitability for deployment in low-resource legal environments.

Traditional Optical Character Recognition (OCR) systems, such as Tesseract~\citep{4376991}, EasyOCR~\citep{easyocr2020}, and PaddleOCR~\citep{li2022ppocrv3attemptsimprovementultra}, follow a multi-stage pipeline involving text detection, segmentation, and \mbox{recognition}. These systems perform reasonably well for printed documents but often fail on handwritten legal content due to limited generalization, poor handwriting support, and a lack of layout-awareness. Moreover, they are not designed to work with low-resource languages like Marathi unless extensively retrained. Our work evaluates these tools on a custom dataset of handwritten Marathi legal documents and investigates their effectiveness when paired with modern MT models like IndicTrans2~\citep{gala2023indictrans2highqualityaccessiblemachine} and Sarvam-1 ~\citep{sarvam1}.

However, OCR-MT pipelines suffer from cascading errors; misrecognized words by OCR adversely affect translation quality. This motivates the use of vLLMs, which can jointly process image and text inputs, reducing dependency on rigid pipeline stages. Models such as Chitrarth~\citep{khan2025chitrarthbridgingvisionlanguage}, Ovis2~\citep{lu2024ovis}, and Maya~\citep{alam2024mayainstructionfinetunedmultilingual} are capable of zero-shot visual reasoning and multilingual output generation. These models offer an attractive alternative for legal text digitization, particularly when working with noisy or incomplete inputs.

The key contributions of this work are:
\begin{itemize}
    \item \emph{OCR Evaluation:} We benchmark Tesseract, EasyOCR, and PaddleOCR on a curated dataset of handwritten Marathi legal documents.
    \item \emph{MT Analysis:} Analyze translation performance using IndicTrans2 and Sarvam-1 models on OCR-extracted text.
    \item \emph{vLLM Benchmarking:} We compare three vision-language models against OCR-MT pipelines, highlighting their ability to perform direct image-to-English translation.
\end{itemize}

\section{Related Work}
\label{sec:related_work}
The literature relevant to our task spans four key areas: (1) advancements in Optical Character Recognition (OCR) systems, especially those capable of processing diverse document layouts; (2) developments in machine translation (MT) for Indian languages; (3) systems for handwritten and printed text extraction in Indian legal documents; and (4) progress in vLLMs that perform tasks such as image captioning, visual QA, and multimodal translation.

% \paragraph{OCR and Document Understanding.}
Recent OCR systems incorporate layout and spatial awareness for enhanced recognition. VISTA-OCR~\citep{hamdi2025vistaocrgenerativeinteractiveend} introduces a generative, layout-aware OCR pipeline using an encoder-decoder framework. olmOCR~\citep{poznanski2025olmocrunlockingtrillionstokens} leverages document-anchoring and fine-tunes the Qwen2-VL-7B-Instruct vLLM to extract structured information. Another approach, +FIRST~\citep{iwana2017judgingbookcover}, improves multimodal transcription by combining OCR outputs from full documents with image features of only the first page.
% 
% \paragraph{OCR in Indian Legal Pipelines.}
Systems such as Nirnayak~\citep{10829440} and related pipelines~\citep{10690289} apply OCR for downstream tasks like translation and summarization in the Indian legal domain. However, their reliance on OCR introduces error propagation, limiting end-to-end accuracy. This motivates the use of vision-language models that can jointly reason over visual and textual modalities.
% 
% \paragraph{Indian Language Machine Translation.}
Several models have advanced MT for Indian languages. Sarvam-1~\citep{sarvam1} is a 2B parameter model optimized for 10 Indian languages and English, demonstrating strong performance through careful data curation. IndicTrans2~\citep{gala2023indictrans2highqualityaccessiblemachine}, based on a transformer encoder-decoder architecture, supports all 22 scheduled languages. Other efforts include Anuvaad~\citep{anuvaad2025} and Nemotron-4-Mini-Hindi-4B~\citep{joshi2025adaptingmultilingualllmslowresource}, the latter trained via continued pre-training for bilingual MT.

% \paragraph{Vision-Language Models.}
Recent vLLMs such as LLaMA 4~\citep{meta2024llama4}, Qwen2.5-VL~\citep{bai2025qwen25vltechnicalreport}, GPT-4V~\citep{yang2023dawnlmmspreliminaryexplorations}, Gemini 2.0 Flash, and PaliGemma 2~\citep{steiner2024paligemma2familyversatile} demonstrate strong multimodal reasoning capabilities but often require high-end resources, making them unsuitable for deployment in low-resource legal infrastructures. Lightweight models like Chitrarth~\citep{khan2025chitrarthbridgingvisionlanguage}, Ovis~\citep{lu2024ovis}, and Maya~\citep{alam2024mayainstructionfinetunedmultilingual} balance efficiency and accuracy and are better suited for real-world deployment in district courts. Our work evaluates such models under zero-shot prompting for handwritten document translation.
% 
% \paragraph{Most Closely Related Work.}
PLATTER~\citep{kasuba2025platterpagelevelhandwrittentext} offers an end-to-end handwritten OCR framework with two-stage processing (handwritten text detection and recognition) and supports 10 Indian languages. TransDocAnalyser~\citep{chakraborty2023transdocanalyserframeworkofflinesemistructured} is tailored to legal F.I.R. documents, combining a FastRCNN+Vision Transformer encoder with a BERT-based decoder fine-tuned for legal vocabulary. Our work differs by focusing on translation from handwritten legal documents and comparing OCR-MT and vLLM approaches under a common benchmark.

\section{Dataset}
\label{sec:dataset}

We utilize a custom dataset curated to reflect real-world legal document scenarios. The dataset comprises approximately 60 scanned PDF documents written in Marathi, collected from authentic legal sources. These documents vary in length and structure, including both single-page and multi-page entries. Each page in the dataset contains handwritten Marathi text, often interspersed with printed text. The documents also include diverse visual elements such as official stamps, seals, signatures, and structured tables, which introduce additional challenges for both text recognition and translation.

To establish a reliable ground truth, the Marathi text was manually translated into English by a team of two native Marathi speakers. These translations were subsequently reviewed by a legal language expert to ensure fidelity, contextual accuracy, and terminological consistency. The resulting high-quality annotations serve as reference outputs for evaluating the performance of both the OCR-MT pipeline and the vLLM approach discussed in this study.

An example page from the dataset is provided in the supplementary materials to illustrate the visual and textual characteristics of the documents. All private and sensitive information has been blurred to preserve confidentiality.

\section{Proposed Methodology}
\label{sec:methodology}

We explore two strategies for translating handwritten Marathi legal documents into English: (1) modular OCR-based translation pipelines, and (2) direct end-to-end translation using Vision Large Language Models (vLLMs). 

% \begin{figure}[t]
%     \centering
%     % Ensure 'figures/figure2.png' exists in your directory
%     \includegraphics[width=\columnwidth]{figures/figure2.png}
%     \caption{Overview of the proposed pipelines. Pipelines 1–6 combine modular OCR models with MT models, while Pipelines 7–9 utilize vLLMs for direct image-to-text translation.}
%     \label{fig:experiments}
% \end{figure}

\begin{figure*}[t]
    \centering
    % --- Row 1 ---
    % Figure 2: Numeric Characters (Contains two images side-by-side)
    \begin{minipage}[t]{0.48\textwidth}
        \centering
        % Adjusted width to fit two images inside this minipage
        \includegraphics[height=0.9cm, keepaspectratio]{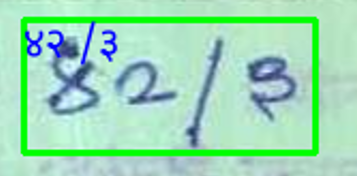}
        \hspace{0.2cm}
        \includegraphics[height=0.9cm, keepaspectratio]{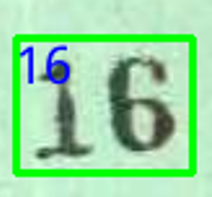}
        \caption{Accurate recognition of printed and handwritten Marathi numeric characters.}
        \label{fig:numeric}
    \end{minipage}
    \hfill
    % Figure 3: Handwritten Dates
    \begin{minipage}[t]{0.48\textwidth}
        \centering
        \includegraphics[height=0.9cm, keepaspectratio]{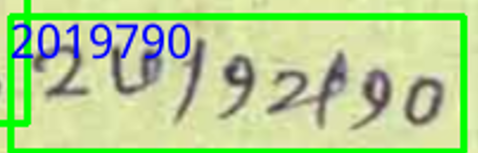}
        \caption{Incorrect extraction of handwritten dates.}
        \label{fig:handwritten-dates}
    \end{minipage}

    \vspace{1em} % Vertical space between rows

    % --- Row 2 ---
    % Figure 4: Printed Text
    \begin{minipage}[t]{0.48\textwidth}
        \centering
        \includegraphics[height=0.9cm, keepaspectratio]{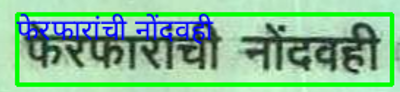}
        \caption{Correct extraction by OCR models for Marathi.}
        \label{fig:printed}
    \end{minipage}
    \hfill
    % Figure 5: Stamp Details
    \begin{minipage}[t]{0.48\textwidth}
        \centering
        \includegraphics[height=0.9cm, keepaspectratio]{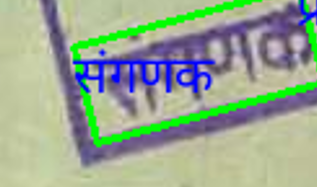}
        \caption{Correct extraction of Marathi stamp details.}
        \label{fig:stamp}
    \end{minipage}
\end{figure*}
\subsection{OCR-MT Pipelines}
\label{subsec:ocr-mt}
We construct six distinct OCR-MT pipelines by combining three OCR tools—Tesseract, EasyOCR, and PaddleOCR—with two state-of-the-art Indian language translation models: Sarvam-1~\citep{sarvam1} and IndicTrans2~\citep{gala2023indictrans2highqualityaccessiblemachine}. Each scanned Marathi document image is first processed by one of the OCR tools to extract textual content. This output is then passed to one of the MT models. This
modular architecture allows a comparative analysis of how
different OCR-MT pairings influence the final translation
quality.
% Specifically, Pipelines 1–2 utilize Tesseract, Pipelines 3–4 utilize EasyOCR, and Pipelines 5–6 utilize PaddleOCR, each paired with Sarvam-1 and IndicTrans2 respectively.

\subsection{Vision-Language Models}
\label{subsec:vllms}
We further evaluate three vision-language models—Chitrarth~\citep{khan2025chitrarthbridgingvisionlanguage}, Maya-8B~\citep{alam2024mayainstructionfinetunedmultilingual}, and Ovis2-34B~\citep{lu2024ovis}—for direct image-to-English translation  without intermediate OCR steps.
These models were tested in a zero-shot setting, guided by manually designed prompts tailored for legal document understanding.
% To optimize performance, we experimented with task-specific prompts for each model; the specific templates used are detailed in Table~\ref{tab:prompts}.
\subsection{Evaluation Protocols}
\label{subsec:evaluation}

We evaluate the OCR outputs using standard fidelity metrics: Character Error Rate (CER)~\citep{k-etal-2025-advocating} and Word Error Rate (WER)~\citep{ali-renals-2018-word}. These quantify the
textual fidelity of Marathi text extracted from the scanned
documents. For translation quality, we conduct human evaluations across the following criteria
\begin{itemize}
    \item \textbf{Fluency:} Grammatical correctness and naturalness of English output.
    \item \textbf{Adequacy:} Degree to which the translation preserves the original meaning.
    \item \textbf{Correctness:} Alignment with gold-standard human \mbox{translations}.
\end{itemize}

All outputs were  evaluated by human annotators fluent in both Marathi and English,  allowing us to identify the most effective and robust pipeline configurations.

\section{Results and Analysis}
\label{sec:results}

%%%%%%%%%%%%%%%%%%%%%%%%%%%%%%%%%%%%%%%%%

OCR models exhibited notably better performance on printed text segments than handwritten ones. While printed segments were recognized with reasonable accuracy (see Figure~\ref{fig:printed}), handwritten content often led to errors, such as omissions, misrecognized characters, or fragmented outputs. Figures~\ref{fig:numeric} and~\ref{fig:handwritten-dates} highlight typical OCR behaviors on handwritten digits and dates, while Figure~\ref{fig:stamp} shows successful stamp extraction. Although EasyOCR consistently outperformed PaddleOCR and Tesseract, it still struggled with inconsistent handwriting styles.

% \begin{table}[t]
%     \centering
%     \small % Allowed: Reduces font to 9pt
%     \setlength{\tabcolsep}{2pt} % Reduces white space between columns to fit more text
%     \begin{tabular}{|c|p{2cm}|p{2.8cm}|p{2cm}|}
%     \hline
%     \textbf{Image} & \textbf{Extracted} & \textbf{English Output} & \textbf{Ground Truth} \\
%     \hline
%     \includegraphics[width=0.9cm, height=0.6cm, keepaspectratio]{figures/img6.png} & Niyam & Rule & Rule \\
%     \hline
%     \includegraphics[width=0.9cm, height=0.6cm, keepaspectratio]{figures/img7.png} & Gaav (Gaon) & Gaon & Village \\
%     \hline
%     \includegraphics[width=0.9cm, height=0.6cm, keepaspectratio]{figures/img7.png} & Gaav (Gaon) & Village & Village \\
%     \hline
%     % The original image was too wide (3.5cm). It is now constrained to the column width.
%     \includegraphics[width=0.9cm, height=0.6cm, keepaspectratio]{figures/img8.png} & Ferfaracha Dinank & Date of Initial Signature or Share Transfer & Date of Change \\
%     \hline
%     \end{tabular}
%     \caption{OCR-MT pipeline: Extracted Marathi text (Latin transliteration) and corresponding English translations compared with human-annotated ground truth.}
%     \label{tab:ocr-translations}
% \end{table}

\begin{table}[t]
    \centering
    \small
    \begin{tabular}{|c|p{1.9cm}|p{2.1cm}|p{1.4cm}|}
    \hline
    \textbf{Image} & \textbf{Extracted} & \textbf{English Output} & \textbf{Ground Truth} \\
    \hline
    \includegraphics[width=0.9cm, height=0.6cm, keepaspectratio]{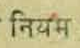} & Niyam & Rule & Rule \\
    \hline
    \includegraphics[width=0.9cm, height=0.6cm, keepaspectratio]{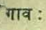} & Gaav (Gaon) & Gaon & Village \\
    \hline
    \includegraphics[width=0.9cm, height=0.6cm, keepaspectratio]{figures/img7.png} & Gaav (Gaon) & Village & Village \\
    \hline
    \includegraphics[width=1.3cm, height=1.3cm, keepaspectratio]{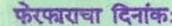} & Ferfaracha Dinank & Date of Share Transfer & Date of Change \\
    \hline
    \end{tabular}
    \caption{OCR-MT pipeline: Extracted Marathi text (Latin transliteration) and corresponding English translations \mbox{compared} with human-annotated ground truth.}
    \label{tab:ocr-translations}
\end{table}

As shown in Table~\ref{tab:ocr-translations}, OCR errors significantly affected translation quality (see Appendix~\ref{app:qualitative}). The pipeline \mbox{sometimes} generated \mbox{incorrect} or misleading translations (e.g., \mbox{mapping} the Marathi word ``Gaav'' (transliterated from ``gaon'', \mbox{meaning} ``village'') to ``Gaon'' instead of ``Village''), or dropped key information. This error propagation across pipeline stages is a core limitation of the OCR-MT approach.

%%%%%%%%%%%%%%%%%%%%%%%%%%%%%%%%%%%%%%%%%
% \begin{figure}[t]
%     \centering
%     \includegraphics[width=0.4\linewidth, height=0.9cm]{figures/img2.png}
%     \includegraphics[width=0.4\linewidth, height=0.9cm]{figures/img3.png}
%     \caption{Accurate recognition of printed and handwritten Marathi numeric characters.}
%     \label{fig:numeric}
% \end{figure}

% \begin{figure}[t]
%     \centering
%     \includegraphics[width=0.5\linewidth, height=0.9cm]{figures/img4.png}
%     \caption{Incorrect extraction of handwritten dates.}
%     \label{fig:handwritten-dates}
% \end{figure}

% \begin{figure}[t]
%     \centering
%     \includegraphics[width=0.5\linewidth, height=0.9cm]{figures/img1.png}
%     \caption{Correct extraction by OCR models for Marathi.}
%     \label{fig:printed}
% \end{figure}

% \begin{figure}[t]
%     \centering
%     \includegraphics[width=0.5\linewidth, height=0.9cm]{figures/img5.png}
%     \caption{Correct extraction of Marathi stamp details.}
%     \label{fig:stamp}
% \end{figure}
%%%%%%%%%%%%%%%%%%%%%%%%%%%%%%%%%%%%%%%%%

Translation models also produced incoherent or mixed-language outputs when fed noisy OCR text. Some translations included trailing untranslated Marathi fragments. Human evaluation revealed loss of legal or factual content, and incomplete or awkward phrasing, particularly for longer sentences with complex legal semantics.

We next evaluated vLLMs, Chitrarth, Maya-8B, and Ovis2 variants, for direct image-to-text translation. Prompt \mbox{engineering} was critical: results improved significantly with detailed instructions. Chitrarth often failed to produce \mbox{coherent} outputs, while Maya-8B showed partial correctness in translation with rich prompts. Ovis2-34B (int4 quantized) and Ovis2-16B also performed inconsistently in zero-shot mode, suggesting a need for domain-specific fine-tuning or prompt-tuning.

Although vLLMs could interpret visual elements and page layout, they often failed to extract precise handwritten content. Generated translations sometimes focused on high-level descriptions of documents rather than their verbatim content. Human annotators rated vLLM outputs based on clarity and fidelity, concluding that while promising, these models currently lack the precision required for legal-grade document translation.

\subsection{Qualitative Comparison of vLLM Outputs}
\label{subsec:qualitative-models}

Table~\ref{table:model_results} (Appendix~\ref{app:qualitative}) presents a qualitative comparison of the \mbox{translated} outputs generated by four vLLMs against the gold-standard human-annotated translation. 
% The selected example illustrates a legally significant passage from a handwritten Marathi document involving a mutation entry and property sale agreement.
% 
The human-annotated \mbox{translation} captures the full context with high fidelity, \mbox{including} masked details of dates, registration \mbox{numbers}, survey \mbox{numbers}, and transaction clauses. In contrast, all four models failed to extract the actual semantics of handwritten content.

\begin{itemize}
    \item \textbf{Chitrarth} produced a hallucinated summary about a meeting, with invented names, dates, and locations that were not present in the original document. This indicates its inability to ground visual input in real text.
    
    \item \textbf{Maya-8B} interpreted the input as a study guide, again showing a lack of alignment with the legal nature of the text. Its output was generic and irrelevant.

    \item \textbf{Ovis2-34B} demonstrated slightly better recognition. It provided partial translations of text such as Marathi ledger headers and dates. However, the content was still largely fabricated or misunderstood.

    \item \textbf{Ovis2-16B} performed relatively better by identifying some legal and financial cues (e.g., account numbers, names, locations). It also translated a few phrases and recognized structural layout, but its output lacked \mbox{completeness} and was partially incoherent.
\end{itemize}

This qualitative example reveals a consistent limitation across current vLLMs: while they are adept at interpreting general visual or structural elements, they fall short of extracting and translating complex handwritten legal content. Moreover, they tend to hallucinate plausible-sounding text when unable to recognize tokens, posing serious risks in high-stakes domains such as law. These findings motivate further efforts in fine-tuning, prompt design, and alignment mechanisms for vLLMs in legal NLP tasks.

\section{Conclusion and Future Scope}
\label{sec:conclusion_future_scope}

This work compared traditional OCR–MT pipelines with end-to-end vision-language models (vLLMs) for translating handwritten Marathi legal documents into English. While OCR-based systems offer modularity and transparency, they suffer from significant error propagation, especially when handling noisy handwritten text. vLLMs provide a unified alternative that bypasses intermediate recognition steps, but current models still struggle with accurately interpreting complex handwriting and domain-specific legal terminology.
Despite these limitations, our results highlight the promising potential of vLLMs for future legal digitization efforts. Their ability to reason over multimodal inputs opens a pathway toward more robust and scalable translation systems that do not rely solely on brittle OCR stages.

\textbf{Future Scope.}  
This study represents an early step in a broader research effort, and several important directions remain:

\begin{itemize}
    \item \textbf{Fine-Tuning on Legal Handwritten Data:} Both OCR and vLLMs require domain-specific adaptation. We plan to fine-tune models on larger annotated datasets of handwritten legal documents.

    \item \textbf{Improved Prompting and Alignment:} Since vLLM \mbox{performance} varies heavily with prompt quality, \mbox{systematic} prompt engineering and instruction alignment will be explored to reduce hallucinations and improve fidelity.

    \item \textbf{Hybrid OCR–vLLM Pipelines:} Combining OCR for structural cues with vLLMs for contextual translation may yield better accuracy than either approach alone.

    \item \textbf{Enhanced Evaluation Protocols:} We will develop \mbox{evaluation} metrics tailored to legal translation, focusing on factual consistency, terminology preservation, and completeness.

    \item \textbf{Model Efficiency and Deployment:} Real-world adoption requires lightweight models that function on district-level compute resources. Techniques such as quantization and distillation will be investigated.

    \item \textbf{Dataset Expansion:} We will continue curating a larger, \mbox{diverse} corpus of handwritten legal documents across states and formats to support model training and \mbox{benchmarking}.
\end{itemize}

Overall, while current models fall short of deployment-ready performance, the insights gained from this work lay a foundation for developing reliable multilingual legal AI systems capable of supporting digital transformation across the Indian judiciary.

\bibliography{aaai2026}
\clearpage
\section*{Appendix}
\label{app:qualitative}
\begin{table}[ht]
\centering
\scriptsize % Reduces font size to fit text in narrow column
\begin{tabular}{|p{0.2\columnwidth}|p{0.75\columnwidth}|}
\hline
\textbf{Model} & \textbf{Translated Text} \\
\hline

\textbf{Ground Truth} & 
On perusal of Mutation Entry No. \textless masked\textgreater ~dated \textless masked\textgreater,~it is recorded that by virtue of the Agreement for Sale dated \textless masked\textgreater ~bearing registration no. \textless masked\textgreater ~registered at the Offices of Sub-registrar of Assurances at \textless masked\textgreater... \textless masked\textgreater ~sold land bearing Survey No. \textless masked\textgreater ~for an area admeasuring (10 m X 81 m) 8 Ares out of 75.1 Ares in favor of \textless masked\textgreater ~and \textless masked\textgreater ~for a consideration of \textless masked\textgreater ~on certain terms and conditions contained therein... \\
\hline

\textbf{Chitrarth} & 
The document is written in Marathi and contains information about the details of a meeting. The text mentions the names of several people, including ``Shri. K. K. Patil", ``Shri. S. S. Patil", and ``Shri. K. K. Patil". The document also contains information about the dates of the meeting... \\
\hline

\textbf{Ovis2-34B (Int4)} & 
The image is a page from an old Marathi ledger... The text at the top reads \textbf{``krushnaavari nidhidand''} which translates to \textit{Krishnaavari Fine Ledger}. \newline
\textbf{Key details:} \newline
\textbf{1) Date:} 23/06/'94 appears multiple times. \newline
- Mentions ``etch.'' and ``57''. \newline
Further text like ``jath.'' suggests amounts paid. Additional details describe transactions and account settlements. \\
\hline

\textbf{Maya-8B} & 
The image displays a handwritten sheet of paper written in the Marathi language. It is a notebook page with various notes... The paper appears to be a study guide or a notebook... \\
\hline

\textbf{Ovis2-16B} &
The document is a financial record in Marathi, dated 27-04. It appears to be an official ledger entry... \newline
Top Section: \newline
1) \textit{kramanchiti nondavane} — Serial Number Entry \newline
Place: Jalgaon \newline
Name \& Account No.: Dattatraya Udaylal Desai... \newline
Amounts: INR 6 lakh = \\
\hline
\end{tabular}
\caption{Qualitative comparison of translations generated by different vision-language models against the \mbox{human-annotated} ground truth.}
\label{table:model_results}
\end{table}

\end{document}